\documentclass[letterpaper, 10 pt, conference]{ieeeconf}  

\IEEEoverridecommandlockouts                              

\usepackage{amsmath} 
\usepackage{amsfonts}
\usepackage{graphicx}
\usepackage{subfigure}
\usepackage{multirow}
\usepackage{cite}
\usepackage{caption}
\usepackage{float} 
\usepackage{subcaption}
\usepackage{authblk}
\usepackage{stfloats}

\title{\LARGE \bf
Neural Ranging Inertial Odometry
}

\author{Si Wang$^{1}$, Bingqi Shen$^{1}$, Fei Wang$^{2}$, Yanjun Cao$^{3}$, Rong Xiong$^{1}$, Yue Wang$^{1*}$
\thanks{This work was supported in part by the National Nature Science Foundation of China under Grant 62373322 and Zhejiang Provincial Natural Science Foundation of China under Grant No. LD24F030001.}%
\thanks{$^{1}$Institute of Cyber-Systems and Control, Zhejiang University, China.}%
\thanks{$^{2}$Beijing Institute of Electronic System Engineering.}%
\thanks{$^{3}$Huzhou Institute of Zhejiang University, Huzhou, China.}%
\thanks{$^{*}$\textit{Corresponding author: Yue Wang.} (E-mail: ywang24@zju.edu.cn)}
}

\begin{document}

\maketitle
\thispagestyle{empty}
\pagestyle{empty}

\begin{abstract}
Ultra-wideband (UWB) has shown promising potential in GPS-denied localization thanks to its lightweight and drift-free characteristics, while the accuracy is limited in real scenarios due to its sensitivity to sensor arrangement and non-Gaussian pattern induced by multi-path or multi-signal interference, which commonly occurs in many typical applications like long tunnels. We introduce a novel neural fusion framework for ranging inertial odometry which involves a graph attention UWB network and a recurrent neural inertial network. Our graph net learns scene-relevant ranging patterns and adapts to any number of anchors or tags, realizing accurate positioning without calibration. Additionally, the integration of least squares and the incorporation of nominal frame enhance overall performance and scalability. The effectiveness and robustness of our methods are validated through extensive experiments on both public and self-collected datasets, spanning indoor, outdoor, and tunnel environments. The results demonstrate the superiority of our proposed IR-ULSG in handling challenging conditions, including scenarios outside the convex envelope and cases where only a single anchor is available.

\end{abstract}

\section{Introduction}
Accurate localization in GPS-denied environments plays a crucial role in robotic applications such as path planning and navigation. Vision-based and LiDAR-based methods can achieve state-of-the-art performance in most cases, but they are constrained by cumulative drift and the degradation of visual or geometric features. Emerging UWB technology has demonstrated promising potential in indoor localization for its scalability, affordable cost and laudable capacity to penetrate occlusion and provide drift-free location\cite{zheng2022optimization,li2023continuous}. 

Despite the occlusion resistance of UWB signals, non-line-of-sight (NLOS) and multi-path conditions still degrade radio signals, bringing in signal delay and ranging bias, which remains an inevitable issue in UWB positioning\cite{li2023continuous, zhao2022finding,zhao2022util}. Apart from that, UWB localization is sensitive to the arrangement\cite{zhao2022finding,wang2020study} and quantity of anchors.
Most approaches depend on at least three non-colinear anchors for positioning\cite{jiang2023efficient,li2018accurate} and the effectiveness will dramatically decrease if the node is outside the convex envelope range of the base stations. Moreover, it should be highlighted that contrary to the assumptions in traditional methods, UWB ranging noise pattern exhibits neither zero-mean nor Gaussian distributions due to surrounding-dependent interference and diffraction which couldn't be parametrically modeled\cite{fishberg2024murp,li2023continuous,hol2009tightly,kok2015indoor,zhao2021learning}. In general, single UWB ranging shows a non-Gaussian pattern and multi-UWB localization is more complex due to multi-signal interference. Conventional methods like multilateration or nonlinear least squares struggle to provide sufficient localization precision without high-demanding calibration. 

The performance of UWB positioning can be improved through the fusion with IMU. Most integration approaches are filtering-based such as Kalman filter\cite{li2018accurate,you2020data,wang2020multiple} and Particle filter\cite{tian2019resetting}. Nevertheless, the accuracy of such solutions is compromised by ignoring higher-order nonlinear terms\cite{kallianpur2013stochastic}, information loss induced by the decoupled fusion manner of various modalities\cite{leutenegger2015keyframe}, and the nature conflicts, that UWB ranging does not follow the zero-mean Gaussian noise assumption.
Furthermore, the existing IMU preintegration does not fully account for the robot's inherent motion patterns and is susceptible to noise or unstable initial values. The integration accuracy can be enhanced through neural-based approaches which is proved by many learned inertial methods\cite{chen2018ionet,liu2020tlio,herath2020ronin,cioffi2023learned}.

In this paper, we harness neural networks to not only model properties that are challenging to capture with classical methods, such as non-Gaussian noise and multi-UWB interference, but also make a more comprehensive fusion of ranging patterns and robot's motion status, which have been ignored or underutilized in the past.
Specifically, a convolutional recurrent neural network (RNN) is used to process inertial sequence and a heterogeneous UWB graph attention network based on \cite{velivckovic2017graph} is proposed to deal with ranging data. 
Graph neural network (GNN) has been widely utilized in various domains including traffic forecasting and recommendation systems\cite{jiang2022graph,wu2022graph} for its flexibility and dynamism. As we all know, the quantity of UWB anchors and tags is not constant due to signal loss or scene transition. The dynamic topological structure and aggregation power of GNN make it ideally suited to address the heterogeneous UWB localization issues.
Beyond modeling ranging attributes, UWB GNN is employed for a neural fusion function to integrate inertial features, ranging features and motion priors provided by IMU neural propagation and least squares (LS) solution. Overall, we propose an IMU-RNN UWB-LS-GNN (\textbf{IR-ULSG}) model for neural ranging inertial odometry and our major contributions are as follows:

\begin{itemize}
  \item We present a neural framework for the fusion of inertial and ranging measurements, realizing a comprehensive fuse of multimodal information and robust, smooth, drift-free localization in challenging scenes.
  \item We propose a graph attention UWB network to learn scene-relevant ranging model and accommodate dynamic fluctuations in the number of anchors or tags, achieving more accurate positioning in real-life scenarios devoid of calibration conditions.
  \item The integration of least squares solution and formulation of nominal frame enhance the performance and scalability of our model. 
  \item We validate our methods in public and self-collected datasets, including indoor, outdoor and tunnel scenarios. The results prove the superiority of our approach under the adverse impacts of being outside the convex envelope area and signal loss that only one anchor exists.
\end{itemize}


\begin{figure*}[bp]
\setlength{\abovecaptionskip}{-0.1cm}
\setlength{\belowcaptionskip}{-0.6cm}
    \centering
    \includegraphics[scale=0.35]{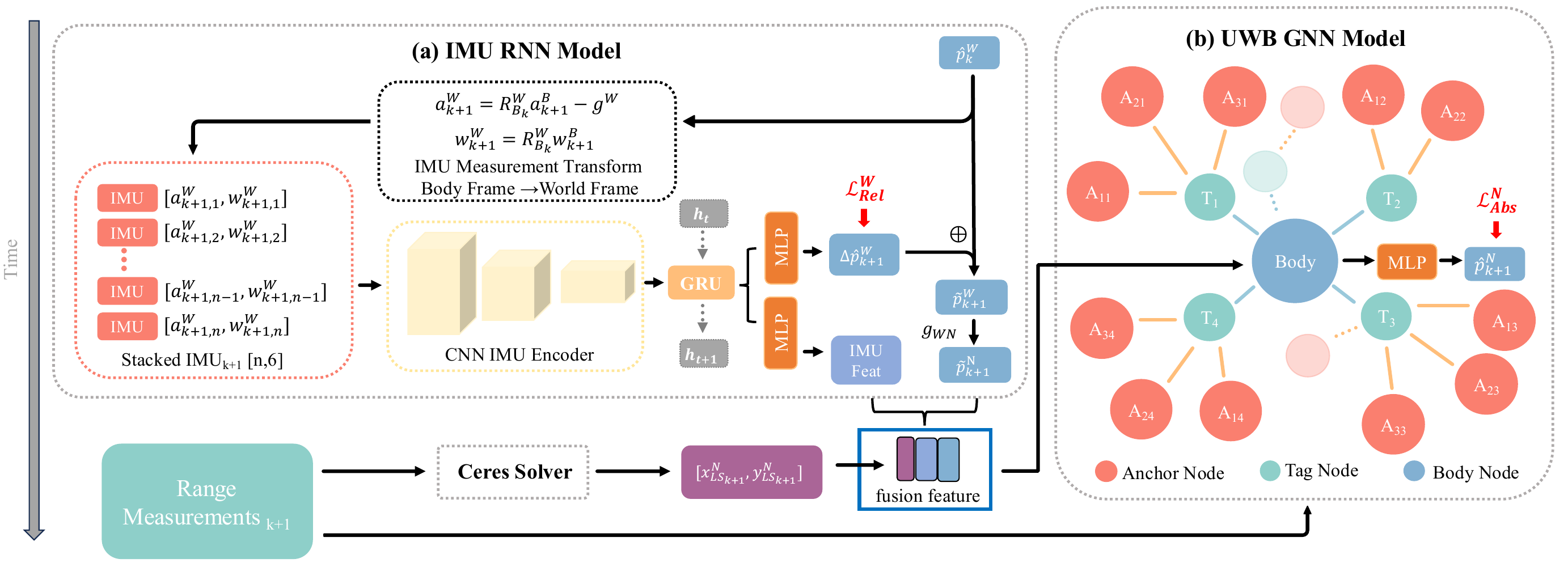}
    \caption{The pipeline of our system}
    \label{fig:pipeline}
\end{figure*}

\section{Related Works}

\subsection{UWB-based Localization} 
UWB is often integrated with other sensors\cite{cao2021vir,nguyen2021viral} to provide drift-free estimation while few works treat UWB as a stand-alone localization solution. Typical UWB-based approaches rely on more than three fixed anchors to determine the location of tags based on trilateration or least squares\cite{shule2020uwb}. However, such methods are sensitive to anchor geometry and environmental impacts such as noise and bias. Ranging bias commonly occurs due to certain factors such as the use of extension cables to place the UWB nodes\cite{nguyen2018robust}, obstruction and reflection of the signals\cite{jiang2023efficient}, inaccurate synchronization, etc. To construct a mild condition for UWB localization, recent 
 studies \cite{zhao2022finding,wang2020study} focus on anchor arrangement to alleviate NLOS effects, a planer pose estimator\cite{jiang2023efficient} calibrate each tag-anchor-pair bias with motion capture system, the work\cite{shalaby2023calibration} propose an antenna-delay calibration procedure for multiple tags. Additionally, the neural network has also been deployed for UWB localization issues. LSTM network is used for positioning in \cite{poulose2020uwb}, investigation\cite{guo2022uwb} realizes better performance in NLOS indoor environment based on genetic annealing and clustering analysis. These works present another avenue to accurately locate, which entails modeling UWB systems by learning. 

\subsection{Learned Inertial Odometry} 
Plagued by unavoidable noise and bias, the continuous integration process is subject to accumulating drifts and ultimately diverges. 
In recent years, there has been a surge in research focusing on learned inertial odometry\cite{chen2018ionet,liu2020tlio,herath2020ronin,cioffi2023learned}. These works highlight the potential of leveraging learned motion priors from IMU sequence, motivating deep-learning-based investigations about neural visual inertial odometry\cite{han2019deepvio,shamwell2019unsupervised,chen2019selective,chen2021rnin}.

\subsection{UWB Inertial Fusion Methods} 
The majority of UWB and IMU fusion methods are filtering-based. 
For example, an Extended Kalman Filter (EKF) typically propagating using IMU measurements and updating the prior with UWB ranges\cite{li2018accurate, cao2020accurate}, an Error State Kalman Filter (ESKF) for 6-DoF state estimation in coal mines\cite{li2020uwb}, an Unscented Kalman Filter (UKF) ignoring the high-order terms for indoor drone localization\cite{you2020data}, all of which belong to the family of Kalman Filters. The accuracy of these approaches is limited by nonlinear terms and non-Gaussian UWB ranging patterns. Beyond Kalman-based methods, Particle Filter (PF)\cite{tian2019resetting} is applied for UWB inertial fusion, which effectively manages non-linear and non-Gaussian problems, but it also faces challenges including particle degeneracy and high computational cost.

\section{Methodology}

Our system overview is shown in Fig. \ref{fig:pipeline}, which mainly consists of an IMU recurrent neural network and a UWB graph neural network. 
IMU RNN model takes in a buffer of IMU measurements and regresses 6-DoF relative transformation as well as inertial features derived from the hidden state of RNN. 
UWB GNN model utilizes the 6-DoF pose prior updated by IMU RNN model and the 2-DoF planer location calculated by Ceres Solver, which are both transformed into nominal frame. With the input of pose prior, inertial features and UWB ranges, GNN finally predicts the 6-DoF pose in nominal frame and transforms it into world frame for the next propagation.

In this section, we first describe the notation used throughout this paper. Second, we clarify how we organize the IMU RNN. Third, we introduce the traditional UWB least squares solved by Ceres, which is also the baseline and part of the pose prior of our graph net. Fourth, we dive into the core UWB GNN. Last, we illustrate the neural fusion of IMU and UWB measurements and the design of loss functions.

\subsection{Notation} 
We denote world frame as $\it W$, whose $\mathit{z}$ axis is aligned with gravity. The UWB nominal frame is $\it N$ and the robot body frame is $\it B$, which is the same as the IMU frame. We use $(\cdot)^{\it W}$ to represent a specific quantity in world frame, likewise for the rest. The position and orientation of $\it B$ with respect to $\it W$ at time $\it t_k$ are written as $\mathbf {t}_{B_k}^{W}\in \mathbb{R}^3$ and $\mathbf {R}_{B_k}^{W}\in \mathbb{R}^{3\times3}$ respectively. The estimated quantities are represented by $\hat{(\cdot)}$ and the prior values are denoted as $\tilde{(\cdot)}$.

\subsection{IMU Recurrent Neural Network} 
Our network consists of the 1D version of CNN, the two-layer-GRU and the MLP. 
Traditional methods need to integrate acceleration data twice to get the relative movement and the integration process is manipulated under the world coordinate frame. To enable neural IMU propagation, we transform the acceleration and angular velocity measurements from body frame $\it B$ to world frame $\it W$ with previous pose, making it easier to learn the gravity-related features and eliminate bias from raw measurements.
\begin{equation}
\setlength{\belowdisplayskip}{0.2pt}
\begin{aligned}
\textbf{\textit{a}} _{k+1}^{W}&=\mathbf {R}_{B_k}^{W}\textbf{\textit{a}} _{k+1}^{B}-\textbf{\textit{g}}^{W} \\
\textbf{\textit{w}} _{k+1}^{W}&=\mathbf {R}_{B_k}^{W}\textbf{\textit{w}} _{k+1}^{B}
\end{aligned}
\end{equation}
where $\textbf{\textit{a}} _{k+1}^{B}$ and $\textbf{\textit{w}} _{k+1}^{B}$ are 
acceleration and angular velocity measurements in the time interval$[\mathit{t}_k , \mathit{t}_{k+1}]$, $\textbf{\textit{g}}^{W}=[0,0,9.8]^T$ is the gravity vector in the world frame. The bias of accelerometer and gyroscope is not explicitly defined in the aforementioned equation, which we assume can be inherently learned in our network.

A batch of IMU data within the time window is fed into 1D-CNN to learn robot motion characteristics. Due to the nature that robot motion being continuous and time-dependent, we employ the two-layer-GRU to fuse the hidden state of the previous frame with the current motion properties. Finally, a MLP is used for regressing the relative pose delta, while another MLP is employed to aggregate the features of the hidden state. Fig. \ref{fig:pipeline}(a) depicts the IMU-RNN (\textbf{IR}) network and the mathematical form is:
\begin{equation}
\setlength{\abovedisplayskip}{3.0pt}
\setlength{\belowdisplayskip}{1.0pt}
\begin{aligned}
&\mathbf{{\bigtriangleup \hat{p}}}_{k+1}^{W},\quad \textbf{\textit{feat}}_{I_{k+1}}=\mathit{f}_{inertial}(\textbf{I} _{k+1}^{W},\textbf{\textit{h}} _{k}) \\
&\textbf{I} _{k+1}^{W}=[(\textbf{\textit{a}} _{k+1,1}^{W},\textbf{\textit{w}} _{k+1,1}^{W}),...,(\textbf{\textit{a}} _{k+1,n}^{W},\textbf{\textit{w}} _{k+1,n}^{W})]
\end{aligned}
\end{equation}
where $\textbf{I} _{k+1}^{W}$ is the stacked IMU measurements. The $\textbf{\textit{h}} _{k}$ is the previous GRU hidden state. The network finally outputs the pose delta $\mathbf{{\bigtriangleup \hat{p}}}_{k+1}^{W}$ and the inertial features $\textbf{\textit{feat}}_{I_{k+1}}$.

\subsection{UWB Least Squares with Ceres Solver}
For scenes where multiple anchors generate range measurements with multiple tags, the UWB ranging model between the i-th tag and the j-th anchor can be defined as follows:
\begin{equation}
\setlength{\abovedisplayskip}{0pt}
\setlength{\belowdisplayskip}{0pt}
\begin{aligned}
\textit{d}_{ij} &= \parallel \mathbf{t}_{t_i}^{W}-\mathbf{t}_{a_j}^{W} \parallel + \, \textit{b}_{ij} + \eta \\
\mathbf{t}_{t_i}^{W} &= \mathbf{R}_{B}^{W} \mathbf{t}_{t_i}^{B} + \mathbf{t}_{B}^{W}
\end{aligned}
\end{equation}
where $\mathbf{t}_{t_i}^{W}, \mathbf{t}_{a_j}^{W} \in \mathbb{R}^3$ refers to the coordinates of the i-th tag and j-th anchor respectively in $\it W$. The $\mathbf{t}_{t_i}^{W}$ is calculated by the tag external parameters $\mathbf{t}_{t_i}^{B}$ and the pose of the robot $\mathbf{R}_{B}^{W}, \mathbf{t}_{B}^{W}$. The $\textit{d}_{ij}, \textit{b}_{ij}$ are range measurements and bias of the corresponding tag-anchor pair. The $\eta$ refers to range measurement noise. Here we omit the subscripts indicating timestamp for simplicity.
We denote the variable to be solved in Ceres as $\boldsymbol{\chi}$:
\begin{equation}
\setlength{\abovedisplayskip}{0.2pt}
\setlength{\belowdisplayskip}{0.2pt}
\begin{aligned}
\boldsymbol{\chi} = [\textit{x}, \textit{y}, \textit{z}, \textit{roll}, \textit{pitch}, \textit{yaw}]^T
\end{aligned}
\end{equation}
where $\textit{x}, \textit{y}, \textit{z}$ refer to the 3D translations and $\textit{roll}, \textit{pitch}, \textit{yaw}$ refer to the 3D rotations.

To find the optimal estimation $\chi^*$, we attempt to minimize the range residuals. Since we lack information on bias in real-world experiments, the residuals are defined as follows:
\begin{equation}
\setlength{\abovedisplayskip}{0.2pt}
\setlength{\belowdisplayskip}{0.2pt}
\begin{aligned}
&\boldsymbol{\chi}^* = argmin(\sum r_{ij}) \\
&r_{ij} = \parallel (\mathbf{R}_{B}^{W} \mathbf{t}_{t_i}^{B} + \mathbf{t}_{B}^{W}) - \mathbf{t}_{a_j}^{W} \parallel^2 - \, \textit{d}_{ij}^2
\end{aligned}
\end{equation}

We use Ceres Solver, as commonly used in the field of SLAM, to solve this least squares problem. The solved 3-DoF position is taken as the baseline for comparison and the 2-DoF planer position is utilized as a pose prior in UWB GNN model.

\subsection{UWB Graph Neural Network}

\textbf{Nominal frame:} Coordinates of the anchors and the robot in world frame $\it W$ vary according to different scenarios. Taking absolute positions as the network input will reduce the generalization ability of the model, also lead to instability in model training. Considering the invariance of distance measurement in Euclidean space, we define a nominal frame $\it N$ according to anchor coordinates within a certain range. The center of the nominal frame is calculated by the center coordinates of the anchor group and the direction is determined by PCA algorithm.
\begin{equation}
\setlength{\abovedisplayskip}{0.8pt}
\setlength{\belowdisplayskip}{0.2pt}
\begin{aligned}
&\mathbf{t}_{N}^{W} = \frac{1}{n} \sum_{j=1}^{n} \mathbf{t}_{a_j}^{W} \\
&\mathbf{R}_{N}^{W} = PCA(\mathbf{t}_{a_1}^{W}, \mathbf{t}_{a_2}^{W}, ..., \mathbf{t}_{a_n}^{W})
\end{aligned}
\end{equation}
where $\mathbf{t}_{N}^{W}$ and $\mathbf{R}_{N}^{W}$ describe the transformation from nominal frame to world frame, $\mathit{n}$ is the number of anchors nearby, $\mathbf{t}_{a_j}^{W}$ is the position of j-th anchor in $\it W$.

By defining the nominal frame, inputs and outputs are confined within a small range. This facilitates the training of the network and simultaneously enhances the generalization ability of the model when applied in multiple scenes.

\begin{table}[b]
    \setlength{\belowcaptionskip}{-0.4cm}
    \setlength{\abovecaptionskip}{-0.03cm}
    \caption{Node Types of GAT}
    \label{table:gat node types}
    \renewcommand\arraystretch{1.2}
    \centering
    \begin{tabular}{cc}
    \hline
        Node Type & Feat \\ \hline
        \textit{Anchor} & $[\mathit{x}_{a}^{N}, \mathit{y}_{a}^{N}, \mathit{z}_{a}^{N}, d]$ \\ 
        \textit{Tag} & $[\mathit{x}_{t}^{B}, \mathit{y}_{t}^{B}, \mathit{z}_{t}^{B}]$ \\ 
        \textit{Body} & $[\mathit{x}_{B}^{N}, \mathit{y}_{B}^{N}, \mathit{z}_{B}^{N}, \mathit{roll}_{B}^{N}, \mathit{pitch}_{B}^{N}, \mathit{yaw}_{B}^{N}]$ \\ \hline
    \end{tabular}
\end{table}

\textbf{UWB GAT model:} 
To effectively extract features of different nodes, we construct a heterogeneous four-layer-GAT model to handle UWB data. There are three different types of nodes in the graph, whose features are listed as TABLE. \ref{table:gat node types}. All the features are stored on the nodes and the edges only represent connections. Features of the \textit{Anchor} node are the coordinates of the anchor in nominal frame and the ranging measurement to a certain tag. If the same anchor generates observations for $\it m$ UWB tags, we duplicate its position $\it m$ times to simplify the topological structure of the graph. For example, the $A_{12}$ node in Fig. \ref{fig:pipeline}(b) contains the 3D coordinates of the 1-st anchor and 1D range between the 1-st anchor and the 2-nd tag. Features of the \textit{Tag} node hold the external parameters of tag in the rigid body frame. The \textit{Body} node takes the pose prior obtained by the previous prediction as node features. A MLP followed by a four-layer-GAT is used to regress 6-DoF pose in nominal frame. The mathematical form of UWB-GNN (\textbf{UG}) model is:

\begin{equation}
\setlength{\abovedisplayskip}{1.0pt}
\setlength{\belowdisplayskip}{0.2pt}
\begin{aligned}
&\mathbf{\hat{p}}_{k+1}^{N} = \mathit{f}_{ranging}(\textbf{U} _{k+1}, \mathbf{\tilde{p}}_{k+1}^{N}) \\
&\textbf{U} _{k+1} = [(\mathbf{t}_{t_1}^{B}...\mathbf{t}_{t_m}^{B}), (\mathbf{t}_{a_1}^{N}... \mathbf{t}_{a_n}^{N}), (\textit{d}_{11}... \, \textit{d}_{mn})]_{k+1} \\ 
&\mathbf{\tilde{p}}_{k+1}^{N} = \mathbf{\hat{p}}_{k}^{N}
\end{aligned}
\end{equation}
where $\mathbf{\hat{p}}_{k+1}^{N}$ refer to the predicted pose in nominal frame. The $\textbf{U} _{k+1}$ represents the range measurements and positions of the tags and anchors, where $\it m$ and $\it n$ respectively refer to the number of tags and anchors. The dependent \textbf{UG} takes the previous estimated pose $\mathbf{\hat{p}}_{k}^{N}$ as the pose prior $\mathbf{\tilde{p}}_{k+1}^{N}$.


\textbf{Integration of LS into GNN:} Since in most scenarios, the layout of anchors is spacious along $\mathit{x}$ and $\mathit{y}$ axes, least squares method provides more stable solutions in these two directions. An intuitive yet innovative idea is to integrate the least squares solution into the network, which we believe can provide a drift-free initial value and accelerate the convergence of the network. We transform the $\mathbf{\tilde{t}}_{LS_{k+1}}^{W}=[\mathit{x}_{LS_{k+1}}^{W}, \mathit{y}_{LS_{k+1}}^{W}]$ solved by Ceres to $\it N$ frame and simply concatenate them with the \textit{Body} node features in the graph. The mathematical form of UWB-LS-GNN (\textbf{ULSG}) is:
\begin{equation}
\setlength{\abovedisplayskip}{0pt}
\setlength{\belowdisplayskip}{0pt}
\begin{aligned}
&\mathbf{\hat{p}}_{k+1}^{N} = \mathit{f}_{ranging}(\textbf{U} _{k+1}, \mathbf{\tilde{p}}_{k+1}^{N}, \mathbf{\tilde{t}}_{LS_{k+1}}^{N}) \\
&\mathbf{\tilde{t}}_{LS_{k+1}}^{N} = \mathit{g}_{WN}(\mathbf{\tilde{t}}_{LS_{k+1}}^{W})
\end{aligned}
\end{equation}
where $\mathit{g}_{WN}$ denotes the transformation from $\it W$ to $\it N$.

\subsection{Neural Fusion of UWB and IMU}
The core module used for fusing UWB and IMU measurements is GNN. Similar to the way we integrate LS into GNN, we concatenate $\textbf{\textit{feat}}_{I_{k+1}}$ with the original \textit{Body} node features. As for the pose delta $\mathbf{{\bigtriangleup \hat{p}}}_{k+1}^{W}$, we use it to update the pose prior for more instant and accurate prior values.
The features $\textbf{\textit{feat}}_{I_{k+1}}$ passed from \textbf{IR} contain the motion state of the robot, uncertainty and noise model of IMU, while the corresponding messages about range measurements are stored in $\textbf{U} _{k+1}$.
Based on the features from IMU and UWB measurements, we believe that GNN is able to determine the weights and fusion methods of these two different modalities. The mathematical form of IMU-RNN UWB-LS-GNN (\textbf{IR-ULSG}) model is:
\begin{equation}
\setlength{\abovedisplayskip}{0.2pt}
\setlength{\belowdisplayskip}{0.2pt}
\begin{aligned}
&\mathbf{\hat{p}}_{k+1}^{N} = \mathit{f}_{ranging}(\textbf{U} _{k+1}, \mathbf{\tilde{p}}_{k+1}^{N}, \mathbf{\tilde{t}}_{LS_{k+1}}^{N}, \textbf{\textit{feat}}_{I_{k+1}}) \\
&\mathbf{\tilde{p}}_{k+1}^{N} = \mathbf{\hat{p}}_{k}^{N} \oplus \mathit{g}_{NW}(\mathbf{{\bigtriangleup \hat{p}}}_{k+1}^{W})
\end{aligned}
\end{equation}
where $\textbf{\textit{feat}}_{I_{k+1}}$ and $\mathbf{{\bigtriangleup \hat{p}}}_{k+1}^{W}$ come from the preceding IMU RNN model. The $\mathbf{{\bigtriangleup \hat{p}}}_{k+1}^{W}$ is transformed to $\it N$ frame by the relation $\mathit{g}_{NW}$ and plus the previous estimated pose $\mathbf{\hat{p}}_{k}^{N}$ to get the pose prior $\mathbf{\tilde{p}}_{k+1}^{N}$ for GNN. The symbol $\oplus$ means direct addition of positions and multiplication of rotation matrices.

\begin{table*}[bp]
\setlength{\belowcaptionskip}{-0.3cm}
\setlength{\abovecaptionskip}{-0.03cm}
\caption{Experiment Datasets}
\label{table:datasets}
\renewcommand\arraystretch{1.2}
    \centering
    \begin{tabular}{ccccccc}
    \hline
        Dataset & Tag/Anchor Num & Loc.Mode & Sensor & GT & Impacts & Movement\\ \hline
        \textit{indoor} & 3/8 & xy & UWB & Motion Capture & Bias/Noise/Multi-Path & $3m\times3m$ \\ 
        \textit{outdoor} & 4/4 & xyz & UWB-IMU & LUIO-SAM & Bias/Noise/Outside Envelope & $40m\times4m$ \\ 
        \textit{tunnel} & 4/36 & xyz & UWB-IMU & LUIO-SAM & Bias/Noise/Multi-Path/NLOS/Anchor Switch & $330m\times60m$ \\ \hline
    \end{tabular}
\end{table*}

\subsection{Loss Functions}
To enable the network to be concerned about the local accuracy while also paying attention to the long-term global accuracy, two different loss functions are applied in our method, both based on the Mean Square Error (MSE) loss. 

To learn the movement during the window between time intervals, we define the relative loss $\mathcal{L}_{Rel}^{W}$ in world frame $\it W$.
To allow the network to eliminate accumulated drift and achieve better global location accuracy, we define the absolute loss $\mathcal{L}_{Abs}^{N}$ in nominal frame $\it N$.
The way we combine relative loss and absolute loss is adjusted by the parameter $\lambda_{Rel}$, which is simply 1.0 in our experiment.
\begin{equation}
\setlength{\abovedisplayskip}{0pt}
\setlength{\belowdisplayskip}{0.2pt}
\begin{aligned}
&\mathcal{L}_{Rel}^{W} = \frac{1}{n} \sum_{k=1}^{n} \parallel \mathbf{{\bigtriangleup p}}_{k}^{W}-\mathbf{{\bigtriangleup \hat{p}}}_{k}^{W} \parallel^2 \\
&\mathcal{L}_{Abs}^{N} = \frac{1}{n} \sum_{k=1}^{n} \parallel \mathbf{p}_{k}^{N}-\mathbf{\hat{p}}_{k}^{N} \parallel^2 \\
&\mathcal{L} = \mathcal{L}_{Rel}^{W}*\lambda_{Rel} + \mathcal{L}_{Abs}^{N}
\end{aligned}
\end{equation}
where $\mathbf{{\bigtriangleup \hat{p}}}_{k}^{W}$ and $\mathbf{\hat{p}}_{k}^{N}$ are pose delta and nominal pose predicted by RNN and GNN respectively, while $\mathbf{{\bigtriangleup p}}_{k}^{W}$ and $\mathbf{p}_{k}^{N}$ are the corresponding ground truth.

\section{Experiment}
In this section, we conduct experiments in multiple scenarios on both public dataset \textit{indoor} and self-collected datasets \textit{outdoor} and \textit{tunnel} to verify the accuracy and robustness of our proposed method compared to traditional methods. In our self-collected datasets, due to the lack of GPS or RTK signals in tunnel environment, the ground truth comes from LUIO-SAM, which is a tightly-coupled LiDAR-UWB-Inertial-Odometry based on LIO-SAM in our previous work. The positions of anchors are solved according to a batch of ranges and estimated robot poses based on least squares. These three scenes suffer from various impacts, characteristics of which are listed in TABLE. \ref{table:datasets}.

\subsection{Experimental Settings and Datasets}
We use a laptop equipped with a 13th Gen Intel Core i9-13900HX CPU and Nvidia RTX 4060 GPU to train and validate our neural network. The \textbf{ULS} solution solved by Ceres serves as our baseline.

\textbf{Indoor:} The open dataset we choose is an indoor planar UWB positioning dataset provided by \cite{jiang2023efficient}, which includes three different dynamic trajectories with various speeds. We use \textit{fast} and \textit{slow} trajectory to train and test on \textit{mid} track. Since the lack of IMU data, the \textbf{IR} module doesn't work. It is worth noting that although \cite{jiang2023efficient} has calibrated range measurements precisely with a motion capture device, we use the raw data in the training and testing process to verify the robustness of the algorithm and its learning ability for unfamiliar environments.

\begin{table}[!ht]
\setlength{\abovecaptionskip}{-0.03cm}
\caption{APE error evaluation under all datasets. Values in bold are the best and in underlined are the second-best.}
\label{table:ape}
\renewcommand\arraystretch{1.2}
\centering
\begin{tabular}{cccccc}
\hline
\multirow{3}{*}{Method} & \multicolumn{5}{c}{Position RMSE (m)}                                                                                                                 \\ \cline{2-6} 
                        & \textit{indoor} & \multicolumn{2}{c}{\textit{outdoor}}                                                                                              & \multicolumn{2}{c}{\textit{tunnel}} \\ \cline{2-6} 
                        & xy     & \begin{tabular}[c]{@{}c@{}}xy\end{tabular} & \begin{tabular}[c]{@{}c@{}}xyz\end{tabular} & xy           & xyz         \\ \hline
ULS            & 0.21   &  \underline{0.52}                                                & 1.06                                                 & \underline{0.28}         & 1.35        \\ 
UG             & \underline{0.10}   & 0.85                                                & 0.88                                                 & 0.55         & 0.72        \\ 
ULSG           & \textbf{0.08}   & -                                                          & -                                                           & -            & -           \\ 
IR-UG          & -      & 0.53                                                & \underline{0.54}                                                 & 0.32         & \underline{0.48}        \\ 
IR-ULSG        & -      & \textbf{0.15}                                                & \textbf{0.19}                                                 & \textbf{0.13}         & \textbf{0.34}        \\ \hline
\end{tabular}
\vspace{-0.2cm}
\end{table}

\begin{figure}[htbp]
    \setlength{\belowcaptionskip}{-0.4cm}
    \centering
    \includegraphics[scale=0.35]{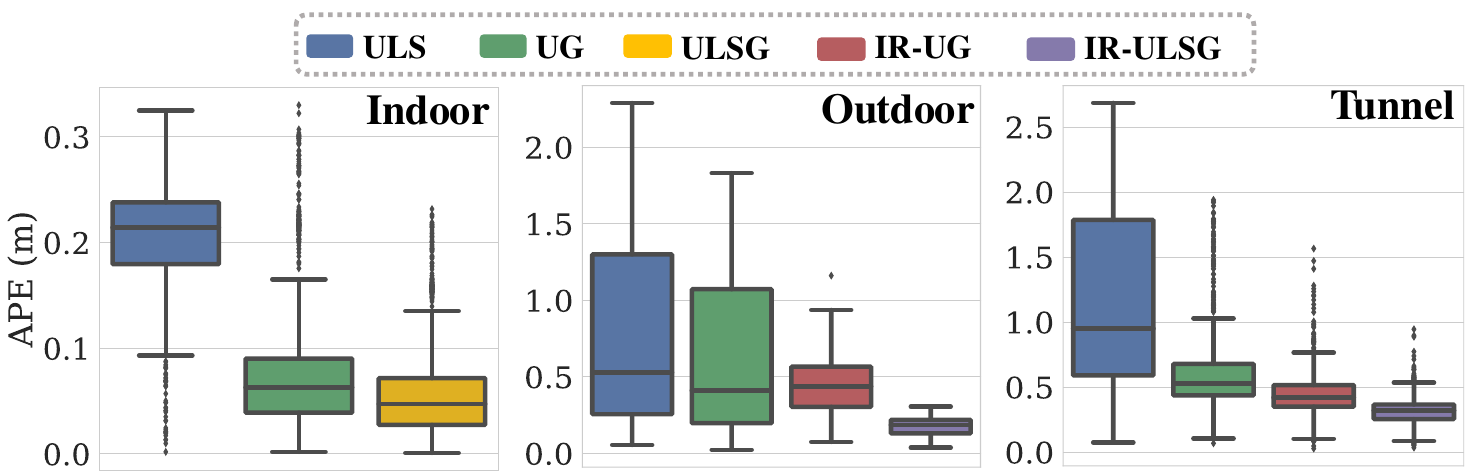}
    \caption{3D APE error box plot under all datasets.}
    \label{fig:box}
\end{figure}

\textbf{Outdoor:} Most UWB localization algorithms can only achieve the utmost performance inside the convex envelope constructed by base stations. In this dataset, the drone flies from outside the envelope to inside.

\textbf{Tunnel:} We also conduct experiments in a 330-meters-long curved tunnel. 36 anchors are arranged in the semi-enclosed environment, in which scene, radio signals will be subject to various adverse effects such as occlusion, reflection, and signal interference between multiple base stations. We only use the ranges within 40m and continuously switch the anchors we accepted as the drone moves.

\subsection{Multi-Scene Localization Accuracy}
Our method is an integration of UWB GNN model, IMU RNN model, and least squares solution provided by Ceres, and each combination of these blocks can provide a pose prediction. In this segment, we will illustrate how each submodule contributes to the whole localization system. We evaluate both 2D and 3D APE positioning error of each submodule by RMSE in TABLE. \ref{table:ape} and the 3D APE error distributions are depicted in Fig. \ref{fig:box}.

\textbf{UWB GNN Model:} \textbf{UG} is designed to learn the UWB ranging and its non-Gaussian noise model. Raw range measurements in \textit{indoor} dataset suffer a mean bias of 18cm, which adversely affects the prediction of LS-based methods. As shown in TABLE. \ref{table:ape} and Fig. \ref{fig:box}, submodules with graph neural networks (\textbf{UG} and \textbf{ULSG}) achieve a notable increase in accuracy over vanilla \textbf{ULS} on \textit{indoor} dataset, which highlights the fact that although we don't explicitly define and supervise the bias term, our network can still learn the bias-related characteristics from training and alleviate its impact without precise but demanding calibration process.

\textbf{Connection with IMU RNN Model:} Aided by IMU RNN model, \textbf{IR-UG} does achieve a higher location accuracy than \textbf{UG}, which validates the role of IMU network. We believe the combination of \textbf{IR} and \textbf{UG} can not only provide a smoother motion estimation but also maintain the effectiveness of positioning when there is large noise or loss in UWB ranging.

\textbf{Integration of Least Squares:} From the contrast between \textbf{UG} and \textbf{ULSG} on \textit{indoor} dataset, \textbf{IR-UG} and \textbf{IR-ULSG} on the other two datasets, it can be inferred that the integration of LS helps to provide more sufficient pose prior. The planer position solved by LS is more globally accurate than that obtained by IMU propagation in some conditions. Consequently, with the approach of neural fusion, the net learns to fuse various features and priors from different sources, which achieves a more favorable pose prediction.

\textbf{Neural Fusion System:} With the combination of UWB GNN model, IMU RNN model and classical least squares method, the overall fusion version \textbf{IR-ULSG} exhibits the utmost precision in localization under all datasets (\textbf{ULSG} on \textit{indoor} dataset for lack of IMU data). Subsequently, we will assess the system's performance to unfavorable disturbances within the environment based on our optimal approach. The trajectories on all datasets estimated by vanilla \textbf{ULS} and ours are presented in Fig. \ref{fig:traj}.

\subsection{Environmental Impacts}
Beyond the bias and noise which are ubiquitous in nearly all environments, UWB localization is also impacted by various detrimental factors such as being outside the convex envelope, multi-path, NLOS or even signal loss. Next we will analyze the system's resilience to interference on our self-collected datasets and the results are shown in TABLE. \ref{table:challenge}.

\begin{table}[]
\setlength{\abovecaptionskip}{-0.03cm}
\setlength{\tabcolsep}{5pt}
\caption{APE error under challenging scenes.}
\label{table:challenge}
\renewcommand\arraystretch{1.2}
\centering
\begin{tabular}{ccccccccc}
\hline
\multirow{4}{*}{Method} & \multicolumn{8}{c}{Position RMSE (m)}                                                                                                                                                \\ \cline{2-9} 
                        & \multicolumn{4}{c}{\begin{tabular}[c]{@{}c@{}}Envelope Coverage\\ (\textit{outdoor})\end{tabular}} & \multicolumn{4}{c}{\begin{tabular}[c]{@{}c@{}}Signal Degradation\\ (\textit{tunnel})\end{tabular}} \\ \cline{2-9} 
                        & \multicolumn{2}{c}{Always In}               & \multicolumn{2}{c}{Out to In}              & \multicolumn{2}{c}{Normal}                  & \multicolumn{2}{c}{Missing}                 \\ \cline{2-9} 
                        & xy                   & xyz                  & xy                   & xyz                 & xy                   & xyz                  & xy                   & xyz                  \\ \hline
ULS                     & 0.33                 & 0.67                 & 0.52                 & 1.06                & 0.28                 & 1.35                 & 0.29                 & 1.28                 \\ 
IR-ULSG                 & \textbf{0.13}        & \textbf{0.15}        & \textbf{0.15}        & \textbf{0.19}       & \textbf{0.13}        & \textbf{0.34}        & \textbf{0.19}        & \textbf{0.27}        \\ \hline
\end{tabular}
\vspace{-0.4cm}
\end{table}

\begin{figure*}[htbp]
\setlength{\abovecaptionskip}{-0.1cm}
\setlength{\belowcaptionskip}{-0.55cm}
\centering
\subfigure[\textit{indoor} scene.]{
\includegraphics[width=.3\textwidth]{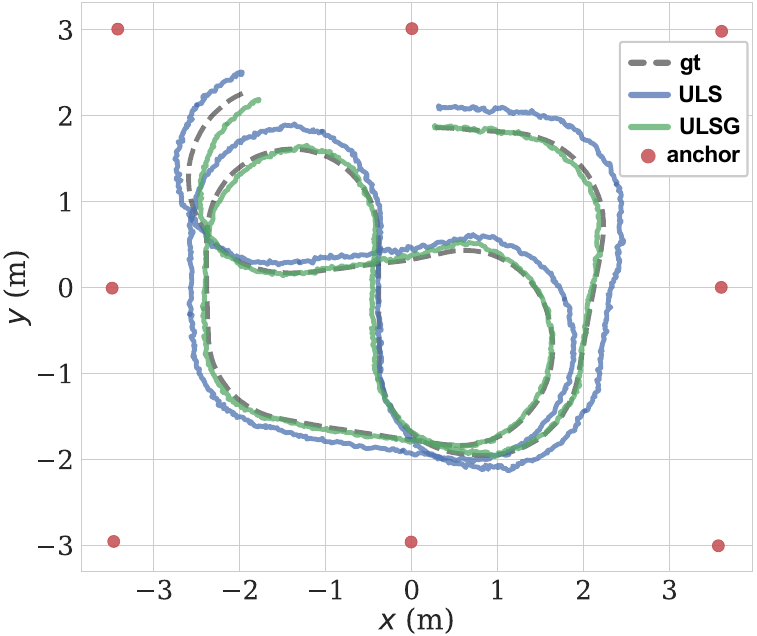}
}
\subfigure[\textit{outdoor} scene.]{
\includegraphics[width=.3\textwidth]{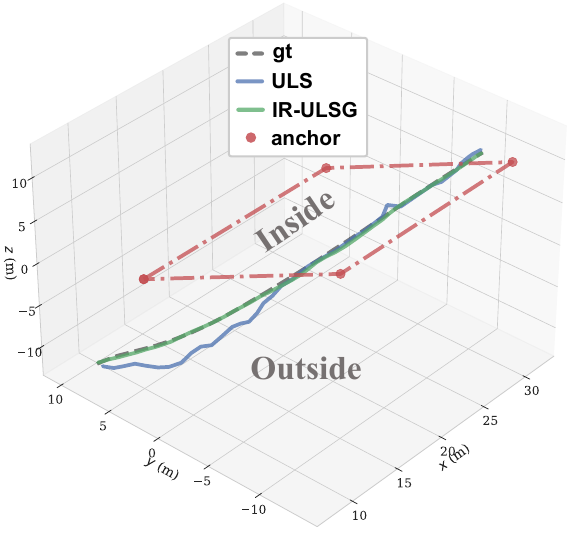}
}
\subfigure[\textit{tunnel} scene.]{
\includegraphics[width=.33\textwidth]{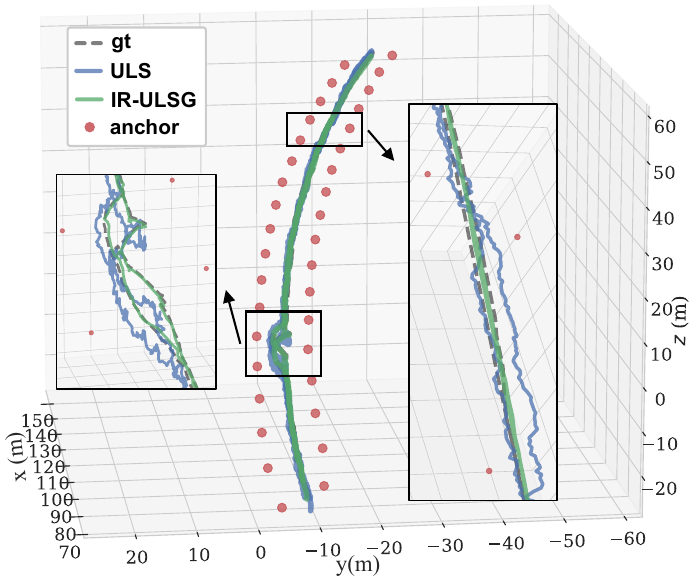}
}
\caption{Trajectories estimated on all datasets. Estimations by vanilla ULS and ours are depicted in blue and green, respectively.}
\label{fig:traj}
\end{figure*}

\begin{figure}[htbp]
    \setlength{\abovecaptionskip}{-0.05cm}
    \setlength{\belowcaptionskip}{-0.6cm}
    \centering
    \includegraphics[width=.4\textwidth]{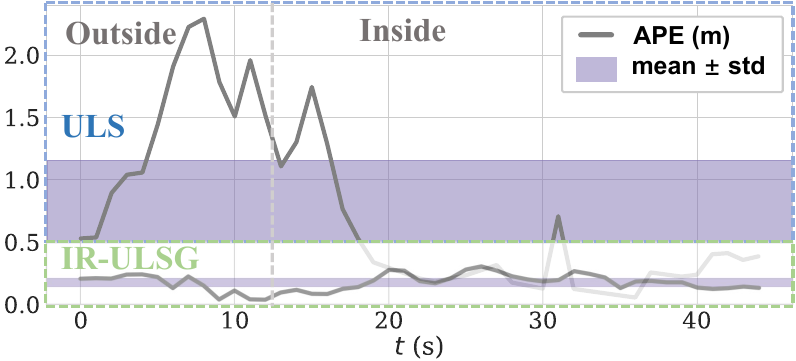}
    \caption{APE error variation on \textit{outdoor} dataset. The blue dashed area represents the error of ULS, and the green dashed box area is the error of IR-ULSG. The boundary of anchor convex envelope is divided by the gray dashed line.}
    \label{fig:outdoor-err}
\end{figure}

\textbf{Convex Envelope Coverage Impacts:} Fig. \ref{fig:traj}(b) displays the predicted trajectories when the UAV flies from outside the anchor envelope area to within the coverage on \textit{outdoor} dataset. The APE error distributions inside and outside the convex hull are statistically demonstrated in Fig. \ref{fig:outdoor-err} and TABLE. \ref{table:challenge}. The comparison indicates when located outside the envelope hull, the performance of \textbf{ULS} drops significantly like many other traditional approaches, while our proposed \textbf{IR-ULSG} maintains high accuracy during the whole process. Our method can still provide stable location where the layout of anchors is limited, which broadens the activity range of robots and the application scenarios of UWB positioning.

\begin{figure}[!ht]
\centering
\subfigure[Trajectories along each axis. The left part is the forth route and the right part is the back route.]{
\includegraphics[width=.47\textwidth]{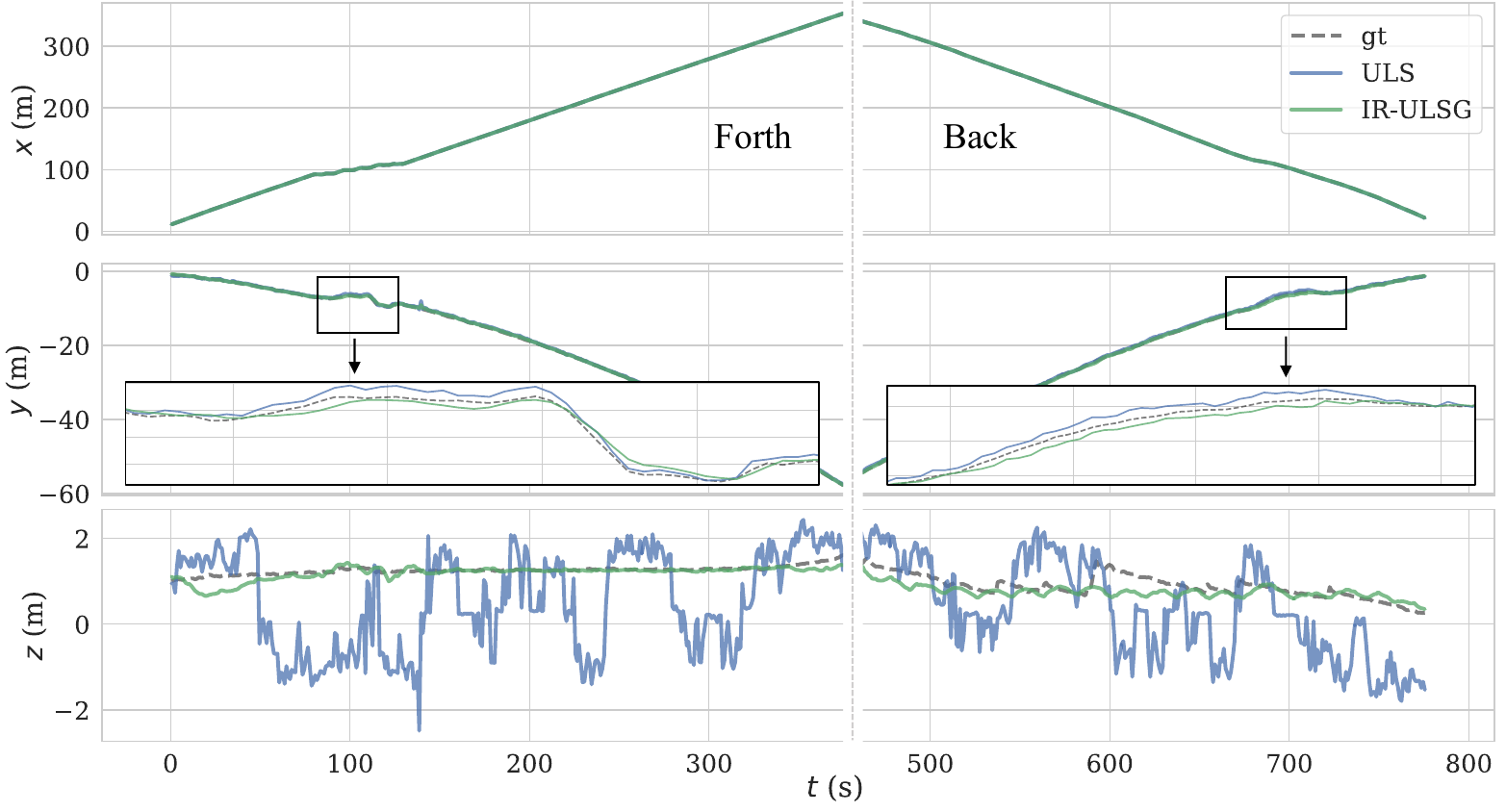}
}
\subfigure[The variation of the number of anchors over time of the forth route. Signal reception is sparse at both ends of the tunnel and dense in the middle of the tunnel in normal circumstances. The gray-bordered region represents the period of anchor signal loss and its corresponding predicted trajectories.]{
\includegraphics[width=.47\textwidth]{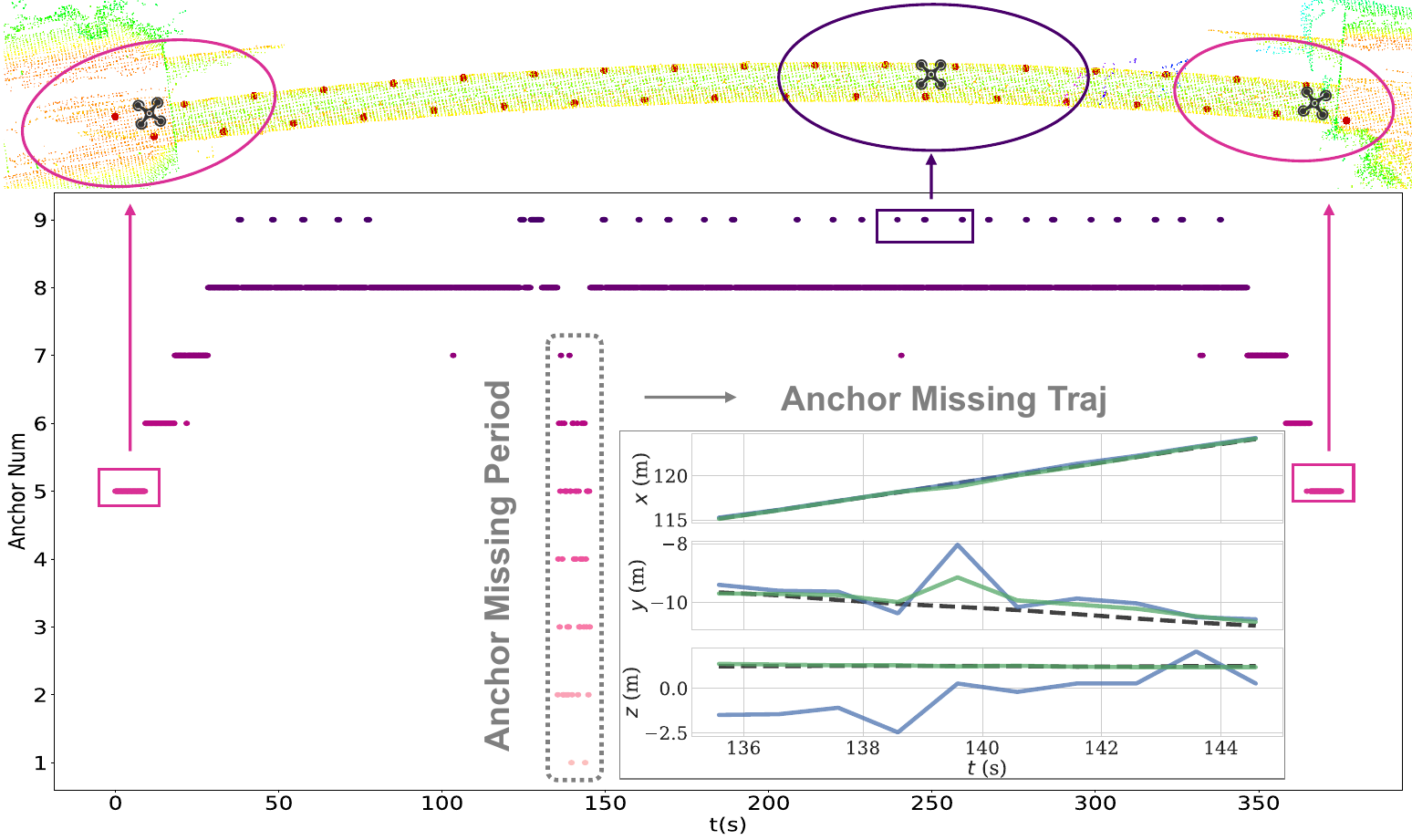}
}
\DeclareGraphicsExtensions.
\caption{Estimations and anchor number variation during anchor missing experiment under \textit{tunnel} dataset.}
\label{fig:tunnel-xyz-anchor}
\end{figure}

\textbf{Anchor Missing Impacts:} To demonstrate the robustness of our method, we allow 30\% of anchors to be randomly lost during the training process on the \textit{tunnel} dataset, and there are 10-second durations where 50\% of the anchors signals missing while testing, including a period of time where only one or two anchors are present.
Fig. \ref{fig:tunnel-xyz-anchor}(a) shows the estimation along each axis. Since the anchors are arranged at the same height and the flight height of the drone is close to that, the location along $\mathit{z}$ axis solved by \textbf{ULS} fluctuates violently. Aided by the neural fusion of inertial features, \textbf{IR-ULSG} ensures higher accuracy and smoothness in 3D positioning. 
As seen in Fig .\ref{fig:tunnel-xyz-anchor}(b), the number of anchors varies within the range of 1 to 9 during the flight. Note that even with only one single anchor functioning, which caused minor perturbations along \textit{y} axis, the deviations were controlled within a narrow scope and rapidly readjusted to normal upon the signal's return.

\begin{figure}[!htbp]
\centering
\subfigure[x axis.]{
\includegraphics[width=.22\textwidth]{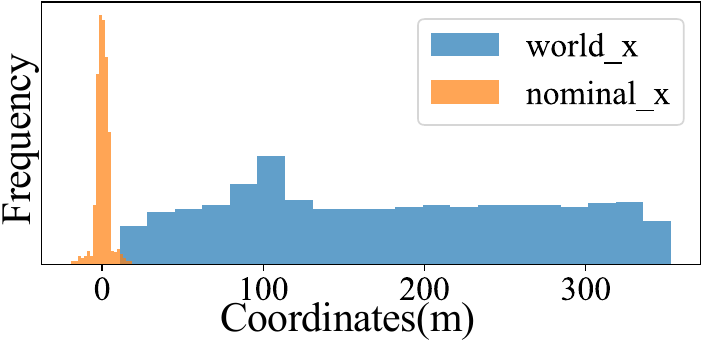}
}
\subfigure[y axis.]{
\includegraphics[width=.22\textwidth]{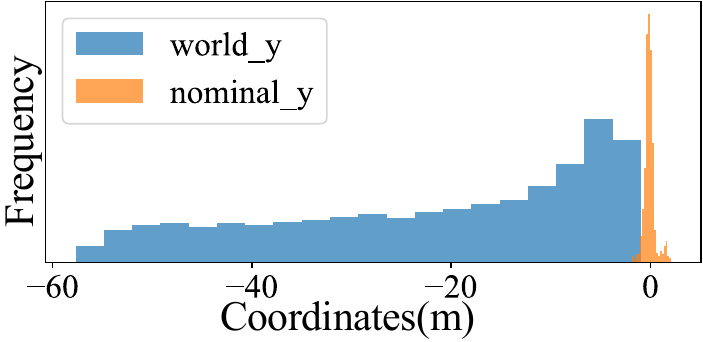}
}
\caption{Coordinates of world frame and nominal frame under \textit{tunnel} dataset.}
\label{fig:nominal}
\end{figure}

\subsection{Nominal Frame}
Last we will discuss the function of nominal frame. In our experiment, the long sequences training with world frame in large scenes will fail due to gradient explosion. As demonstrated in Fig. \ref{fig:nominal}, the nominal coordinates can still be restricted within a small range even in large-scale scenes, enhancing the system's generalization and scalability.

\section{Conclusions}
In this work, we propose a neural odometry framework using ultra-wideband-inertial measurements. GNN is exploited for UWB modeling and the key component to fuse IMU RNN model and LS solution. The neural fusion structure allows more comprehensive use of multimodal information, achieving smooth and reliable estimation. Experiments in challenging scenes demonstrate the robustness and accuracy of our approach coping with detrimental influences. Based on the neural architecture, we believe the performance can be further improved by substituting the LS solution solved by Ceres with a higher precision algorithm or incorporating ancillary information, such as received signal strength (RSS) and channel impulse response (CIR).


\bibliographystyle{IEEEtran}
\bibliography{Mybib}

@article{zheng2022optimization,
  title={An optimization-based UWB-IMU fusion framework for UGV},
  author={Zheng, Shuaikang and Li, Zhitian and Liu, Yunfei and Zhang, Haifeng and Zou, Xudong},
  journal={IEEE Sensors Journal},
  volume={22},
  number={5},
  pages={4369--4377},
  year={2022},
  publisher={IEEE}
}

@article{li2023continuous,
  title={Continuous-time ultra-wideband-inertial fusion},
  author={Li, Kailai and Cao, Ziyu and Hanebeck, Uwe D},
  journal={IEEE Robotics and Automation Letters},
  volume={8},
  number={7},
  pages={4338--4345},
  year={2023},
  publisher={IEEE}
}

@article{zhao2022finding,
  title={Finding the right place: Sensor placement for UWB time difference of arrival localization in cluttered indoor environments},
  author={Zhao, Wenda and Goudar, Abhishek and Schoellig, Angela P},
  journal={IEEE Robotics and Automation Letters},
  volume={7},
  number={3},
  pages={6075--6082},
  year={2022},
  publisher={IEEE}
}

@article{zhao2022util,
  title={UTIL: An ultra-wideband time-difference-of-arrival indoor localization dataset},
  author={Zhao, Wenda and Goudar, Abhishek and Qiao, Xinyuan and Schoellig, Angela P},
  journal={The International Journal of Robotics Research},
  pages={02783649241230640},
  year={2022},
  publisher={SAGE Publications Sage UK: London, England}
}

@article{fishberg2024murp,
  title={MURP: Multi-Agent Ultra-Wideband Relative Pose Estimation with Constrained Communications in 3D Environments},
  author={Fishberg, Andrew and Quiter, Brian and How, Jonathan P},
  journal={IEEE Robotics and Automation Letters},
  year={2024},
  publisher={IEEE}
}

@article{wang2020study,
  title={A study on the optimization nodes arrangement in UWB localization},
  author={Wang, Shijia and Wang, Shibo and Liu, Wanli and Tian, Ye},
  journal={Measurement},
  volume={163},
  pages={108056},
  year={2020},
  publisher={Elsevier}
}

@inproceedings{hol2009tightly,
  title={Tightly coupled UWB/IMU pose estimation},
  author={Hol, Jeroen D and Dijkstra, Fred and Luinge, Henk and Schon, Thomas B},
  booktitle={2009 IEEE international conference on ultra-wideband},
  pages={688--692},
  year={2009},
  organization={IEEE}
}

@article{kok2015indoor,
  title={Indoor positioning using ultrawideband and inertial measurements},
  author={Kok, Manon and Hol, Jeroen D and Sch{\"o}n, Thomas B},
  journal={IEEE Transactions on Vehicular Technology},
  volume={64},
  number={4},
  pages={1293--1303},
  year={2015},
  publisher={IEEE}
}

@article{zhao2021learning,
  title={Learning-based bias correction for time difference of arrival ultra-wideband localization of resource-constrained mobile robots},
  author={Zhao, Wenda and Panerati, Jacopo and Schoellig, Angela P},
  journal={IEEE Robotics and Automation Letters},
  volume={6},
  number={2},
  pages={3639--3646},
  year={2021},
  publisher={IEEE}
}

@article{li2020uwb,
  title={UWB-based localization system aided with inertial sensor for underground coal mine applications},
  author={Li, Meng-Gang and Zhu, Hua and You, Shao-Ze and Tang, Chao-Quan},
  journal={IEEE Sensors Journal},
  volume={20},
  number={12},
  pages={6652--6669},
  year={2020},
  publisher={IEEE}
}

@inproceedings{nguyen2018robust,
  title={Robust target-relative localization with ultra-wideband ranging and communication},
  author={Nguyen, Thien-Minh and Zaini, Abdul Hanif and Wang, Chen and Guo, Kexin and Xie, Lihua},
  booktitle={2018 IEEE international conference on robotics and automation (ICRA)},
  pages={2312--2319},
  year={2018},
  organization={IEEE}
}

@article{jiang2022graph,
  title={Graph neural network for traffic forecasting: A survey},
  author={Jiang, Weiwei and Luo, Jiayun},
  journal={Expert systems with applications},
  volume={207},
  pages={117921},
  year={2022},
  publisher={Elsevier}
}

@article{wu2022graph,
  title={Graph neural networks in recommender systems: a survey},
  author={Wu, Shiwen and Sun, Fei and Zhang, Wentao and Xie, Xu and Cui, Bin},
  journal={ACM Computing Surveys},
  volume={55},
  number={5},
  pages={1--37},
  year={2022},
  publisher={ACM New York, NY}
}

@article{velivckovic2017graph,
  title={Graph attention networks},
  author={Veli{\v{c}}kovi{\'c}, Petar and Cucurull, Guillem and Casanova, Arantxa and Romero, Adriana and Lio, Pietro and Bengio, Yoshua},
  journal={arXiv preprint arXiv:1710.10903},
  year={2017}
}

@inproceedings{chen2018ionet,
  title={Ionet: Learning to cure the curse of drift in inertial odometry},
  author={Chen, Changhao and Lu, Xiaoxuan and Markham, Andrew and Trigoni, Niki},
  booktitle={Proceedings of the AAAI Conference on Artificial Intelligence},
  volume={32},
  number={1},
  year={2018}
}

@article{liu2020tlio,
  title={Tlio: Tight learned inertial odometry},
  author={Liu, Wenxin and Caruso, David and Ilg, Eddy and Dong, Jing and Mourikis, Anastasios I and Daniilidis, Kostas and Kumar, Vijay and Engel, Jakob},
  journal={IEEE Robotics and Automation Letters},
  volume={5},
  number={4},
  pages={5653--5660},
  year={2020},
  publisher={IEEE}
}

@inproceedings{herath2020ronin,
  title={Ronin: Robust neural inertial navigation in the wild: Benchmark, evaluations, \& new methods},
  author={Herath, Sachini and Yan, Hang and Furukawa, Yasutaka},
  booktitle={2020 IEEE international conference on robotics and automation (ICRA)},
  pages={3146--3152},
  year={2020},
  organization={IEEE}
}

@article{cioffi2023learned,
  title={Learned inertial odometry for autonomous drone racing},
  author={Cioffi, Giovanni and Bauersfeld, Leonard and Kaufmann, Elia and Scaramuzza, Davide},
  journal={IEEE Robotics and Automation Letters},
  volume={8},
  number={5},
  pages={2684--2691},
  year={2023},
  publisher={IEEE}
}

@inproceedings{han2019deepvio,
  title={Deepvio: Self-supervised deep learning of monocular visual inertial odometry using 3d geometric constraints},
  author={Han, Liming and Lin, Yimin and Du, Guoguang and Lian, Shiguo},
  booktitle={2019 IEEE/RSJ International Conference on Intelligent Robots and Systems (IROS)},
  pages={6906--6913},
  year={2019},
  organization={IEEE}
}

@article{shamwell2019unsupervised,
  title={Unsupervised deep visual-inertial odometry with online error correction for RGB-D imagery},
  author={Shamwell, E Jared and Lindgren, Kyle and Leung, Sarah and Nothwang, William D},
  journal={IEEE transactions on pattern analysis and machine intelligence},
  volume={42},
  number={10},
  pages={2478--2493},
  year={2019},
  publisher={IEEE}
}

@inproceedings{chen2019selective,
  title={Selective sensor fusion for neural visual-inertial odometry},
  author={Chen, Changhao and Rosa, Stefano and Miao, Yishu and Lu, Chris Xiaoxuan and Wu, Wei and Markham, Andrew and Trigoni, Niki},
  booktitle={Proceedings of the IEEE/CVF Conference on Computer Vision and Pattern Recognition},
  pages={10542--10551},
  year={2019}
}

@inproceedings{chen2021rnin,
  title={RNIN-VIO: Robust neural inertial navigation aided visual-inertial odometry in challenging scenes},
  author={Chen, Danpeng and Wang, Nan and Xu, Runsen and Xie, Weijian and Bao, Hujun and Zhang, Guofeng},
  booktitle={2021 IEEE International Symposium on Mixed and Augmented Reality (ISMAR)},
  pages={275--283},
  year={2021},
  organization={IEEE}
}

@inproceedings{jiang2023efficient,
  title={Efficient planar pose estimation via UWB measurements},
  author={Jiang, Haodong and Wang, Wentao and Shen, Yuan and Li, Xinghan and Ren, Xiaoqiang and Mu, Biqiang and Wu, Junfeng},
  booktitle={2023 IEEE International Conference on Robotics and Automation (ICRA)},
  pages={1954--1960},
  year={2023},
  organization={IEEE}
}

@inproceedings{li2018accurate,
  title={Accurate 3D localization for MAV swarms by UWB and IMU fusion},
  author={Li, Jiaxin and Bi, Yingcai and Li, Kun and Wang, Kangli and Lin, Feng and Chen, Ben M},
  booktitle={2018 IEEE 14th International Conference on Control and Automation (ICCA)},
  pages={100--105},
  year={2018},
  organization={IEEE}
}

@article{you2020data,
  title={Data fusion of UWB and IMU based on unscented Kalman filter for indoor localization of quadrotor UAV},
  author={You, Weide and Li, Fanbiao and Liao, Liqing and Huang, Meili},
  journal={Ieee Access},
  volume={8},
  pages={64971--64981},
  year={2020},
  publisher={IEEE}
}

@article{wang2020multiple,
  title={Multiple-vehicle localization using maximum likelihood Kalman filtering and ultra-wideband signals},
  author={Wang, Wenxu and Marelli, Damian and Fu, Minyue},
  journal={IEEE Sensors Journal},
  volume={21},
  number={4},
  pages={4949--4956},
  year={2020},
  publisher={IEEE}
}

@article{tian2019resetting,
  title={A resetting approach for INS and UWB sensor fusion using particle filter for pedestrian tracking},
  author={Tian, Qinglin and Kevin, I and Wang, Kai and Salcic, Zoran},
  journal={IEEE Transactions on Instrumentation and Measurement},
  volume={69},
  number={8},
  pages={5914--5921},
  year={2019},
  publisher={IEEE}
}

@book{kallianpur2013stochastic,
  title={Stochastic filtering theory},
  author={Kallianpur, Gopinath},
  volume={13},
  year={2013},
  publisher={Springer Science \& Business Media}
}

@article{leutenegger2015keyframe,
  title={Keyframe-based visual--inertial odometry using nonlinear optimization},
  author={Leutenegger, Stefan and Lynen, Simon and Bosse, Michael and Siegwart, Roland and Furgale, Paul},
  journal={The International Journal of Robotics Research},
  volume={34},
  number={3},
  pages={314--334},
  year={2015},
  publisher={SAGE Publications Sage UK: London, England}
}

@article{cao2021vir,
  title={VIR-SLAM: Visual, inertial, and ranging SLAM for single and multi-robot systems},
  author={Cao, Yanjun and Beltrame, Giovanni},
  journal={Autonomous Robots},
  volume={45},
  number={6},
  pages={905--917},
  year={2021},
  publisher={Springer}
}

@article{nguyen2021viral,
  title={Viral slam: Tightly coupled camera-imu-uwb-lidar slam},
  author={Nguyen, Thien-Minh and Yuan, Shenghai and Cao, Muqing and Nguyen, Thien Hoang and Xie, Lihua},
  journal={arXiv preprint arXiv:2105.03296},
  year={2021}
}

@article{shule2020uwb,
  title={Uwb-based localization for multi-uav systems and collaborative heterogeneous multi-robot systems},
  author={Shule, Wang and Almansa, Carmen Mart{\'\i}nez and Queralta, Jorge Pe{\~n}a and Zou, Zhuo and Westerlund, Tomi},
  journal={Procedia Computer Science},
  volume={175},
  pages={357--364},
  year={2020},
  publisher={Elsevier}
}

@inproceedings{shalaby2023calibration,
  title={Calibration and uncertainty characterization for ultra-wideband two-way-ranging measurements},
  author={Shalaby, Mohammed Ayman and Cossette, Charles Champagne and Forbes, James Richard and Le Ny, Jerome},
  booktitle={2023 IEEE International Conference on Robotics and Automation (ICRA)},
  pages={4128--4134},
  year={2023},
  organization={IEEE}
}

@article{poulose2020uwb,
  title={UWB indoor localization using deep learning LSTM networks},
  author={Poulose, Alwin and Han, Dong Seog},
  journal={Applied Sciences},
  volume={10},
  number={18},
  pages={6290},
  year={2020},
  publisher={MDPI}
}

@article{guo2022uwb,
  title={UWB indoor positioning optimization algorithm based on genetic annealing and clustering analysis},
  author={Guo, Hua and Li, Mengqi and Zhang, Xuejing and Gao, Xiaotian and Liu, Qian},
  journal={Frontiers in Neurorobotics},
  volume={16},
  pages={715440},
  year={2022},
  publisher={Frontiers Media SA}
}

@inproceedings{cao2020accurate,
  title={Accurate position tracking with a single UWB anchor},
  author={Cao, Yanjun and Yang, Chenhao and Li, Rui and Knoll, Alois and Beltrame, Giovanni},
  booktitle={2020 IEEE international conference on robotics and automation (ICRA)},
  pages={2344--2350},
  year={2020},
  organization={IEEE}
}

\end{document}